\documentclass[lettersize,journal]{IEEEtran}
\usepackage{amsmath,amsfonts}
\usepackage{graphicx}
\usepackage{cite}
\usepackage{booktabs}
\usepackage{multirow}
\usepackage{url}

\begin{document}

\title{Radar-Guided Camera Verification for Automatic Emergency Braking: Rethinking Object Detection in Radar--Camera Fusion}
\author{%
Ram Charan Akula,
Sivanathan Kandhasamy,
and Manikandan Ganesan%
\thanks{Manuscript received Month XX, 2026; revised Month XX, 2026.}%
\thanks{Ram Charan and Sivanathan Kandhasamy are with the Autonomous Systems Lab, Department of Mechatronics Engineering, SRM Institute of Science and Technology, Kattankulathur, Chennai 603203, India (e-mail: ramcharan@srmist.edu.in; sivanak@srmist.edu.in).}%
\thanks{Manikandan Ganesan is with Centre for Electric Mobility, Department of EEE, SRM Institute of Science and Technology, Kattankulathur, Chennai 603203, India}%
\thanks{Corresponding author: Sivanathan Kandhasamy (e-mail: sivanak@srmist.edu.in).}%
}

\maketitle

\begin{abstract}
Radar--camera fusion is widely used in Automatic Emergency Braking (AEB) systems because radar provides reliable range and velocity measurements while cameras provide a proper  visual confirmation of the objects . Most of the deployed systems perform this confirmation using computationally intensive object detectors. However, if the  radar has already localized a target, the camera may only need to verify the obstacle's presence rather than solving a full problem by identifying the object. 
Our work proposes a radar-scoped edge-density gate that performs obstacle verification within radar-guided image regions of interest. This method requires no training data, model weights, or GPU acceleration and was integrated into a complete radar--camera fusion AEB system with brake-by-wire actuation. Evaluated on a real instrumented vehicle across 72 driving sessions and 131,603 camera frames, the proposed approach reduced the camera search space by up to 98.7\%, achieved a mean processing latency of 0.121 ms per ROI, an AUC of 0.898, and a recall of 0.994. Across 33 staged threat scenarios, the complete AEB system recorded zero missed brake events.\end{abstract}


\section{Introduction}

The Automatic Emergency Braking (AEB) system relies heavily on radar--camera fusion to make reliable braking decisions. Radar provides robust measurements of range and relative velocity, while cameras provide visual information for obstacle confirmation \cite{who2018roadsafety,nabati2021centerfusion,zhou2023bridging,wang2023review,lin2023embeddedfusion}. As a result, radar--camera fusion has become one of the most widely adopted perception architectures in the modern AEB systems \cite{jahromi2019hybridfusion,baumann2024cr3dt,wang2020multisensor,shi2024radarcamerasurvey, wu2024radarvisionsurvey,wei2022mmwave}.

In most of the fusion architectures, the camera performs a semantic object detection using deep learning models \cite{nabati2021centerfusion,zhou2023bridging,wang2023review,lin2023embeddedfusion,baumann2024cr3dt, ganesan2024motionreview}, This approach has significantly improved perception performance, but it also increases the computational complexity and hardware requirements of the system. More importantly, it assumes that the camera must solve two problems: locating the obstacle within the scene and identifying what the obstacle is.

In a radar-led AEB pipeline, the first problem has already been solved. In a radar-led AEB pipeline, radar detects, tracks, and localizes the target before the camera is used for verification. Once the target has been projected into the image plane, the camera no longer needs to search the entire frame for potential obstacles. It only needs to evaluate a small radar-guided region of interest associated with the tracked target.

The observation raises a second question. If radar has already provided localization, does the camera still need to perform full object recognition? For an AEB system, the braking decision depends primarily on whether an obstacle exists in the vehicle's path, not necessarily on assigning a semantic label to that obstacle. This changes the role of the camera from detector to verifier.

\begin{figure*}[t]
\centering
\includegraphics[width=\textwidth]{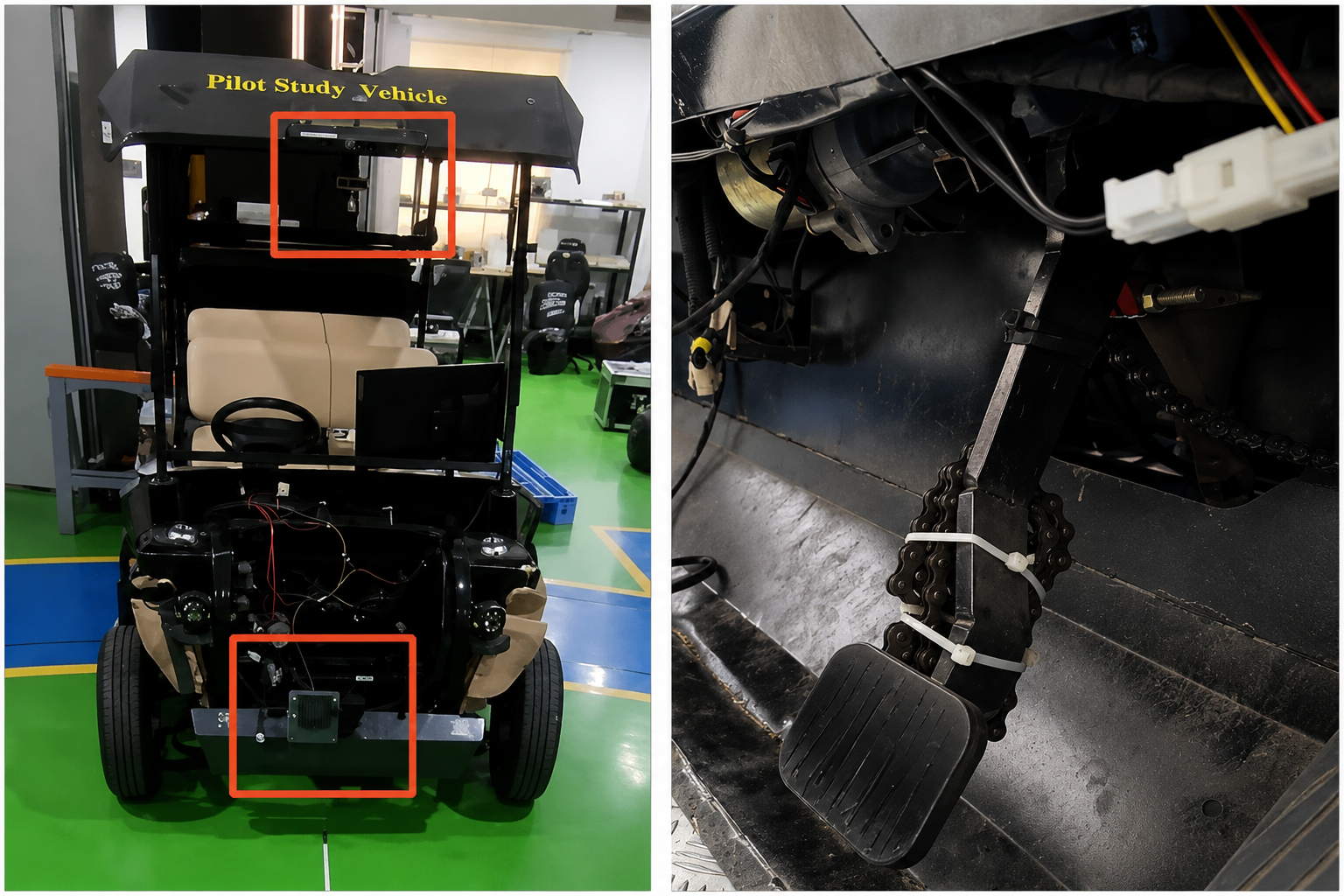}
\caption{On the left side we have the instrumented test vehicle---the upper orange box marks the RealSense D435 camera, lower box marks the ARS408-21 radar. Right: Chain-sprocket brake-by-wire actuator pulling the brake pedal.}
\label{test_vehicle}
\end{figure*}

To investigate this idea, we developed a radar-scoped edge-density verification gate operating on radar-projected image regions. The proposed approach requires no training data, model weights, or GPU acceleration and was integrated into a complete radar--camera fusion AEB system comprising radar processing, Kalman-filter-based target tracking \cite{kalman1960filter}, camera projection, obstacle verification using edge extraction techniques \cite{canny1986edge}, and brake-by-wire actuation. The system was deployed on a real vehicle using an ARS408-21 automotive radar \cite{continental2021ars408} and an Intel RealSense D435 camera \cite{intel2020realsense}.

The principal contributions of this work are as follows:

\begin{enumerate}
    \item A radar-guided edge-density gate that uses the camera for obstacle verification rather than object detection.
    \item A complete radar--camera fusion AEB system implemented and tested on a real vehicle with brake-by-wire actuation.
    \item An experimental evaluation of latency, verification performance, and system behaviour using real-world driving data.
    \item A study of whether lightweight camera verification can replace detector-based confirmation in resource-constrained AEB systems.
\end{enumerate}

\section{Related Work}

Radar--camera fusion is widely used in autonomous driving because radar provides reliable range and velocity measurements while cameras provide visual information \cite{wang2023review,wang2020multisensor, shi2024radarcamerasurvey, wu2024radarvisionsurvey,wei2022mmwave}. Most of the recent research has focused primarily on improving perception performance through increasingly sophisticated fusion architectures and learned detection models \cite{ganesan2024motionreview,shi2024radarcamerasurvey,wu2024radarvisionsurvey,wei2022mmwave}. Representative examples include CenterFusion \cite{nabati2021centerfusion}, feature-level fusion approaches \cite{zhou2023bridging}, and camera--radar detection and tracking frameworks such as CR3DT \cite{baumann2024cr3dt}. Despite differences in implementation, these systems generally use the camera as an object detector and rely on learned models for obstacle recognition and scene understanding \cite{nabati2021centerfusion,zhou2023bridging,baumann2024cr3dt, ganesan2024motionreview}.

\begin{figure*}[t]
\centering
 \includegraphics[width=\textwidth]{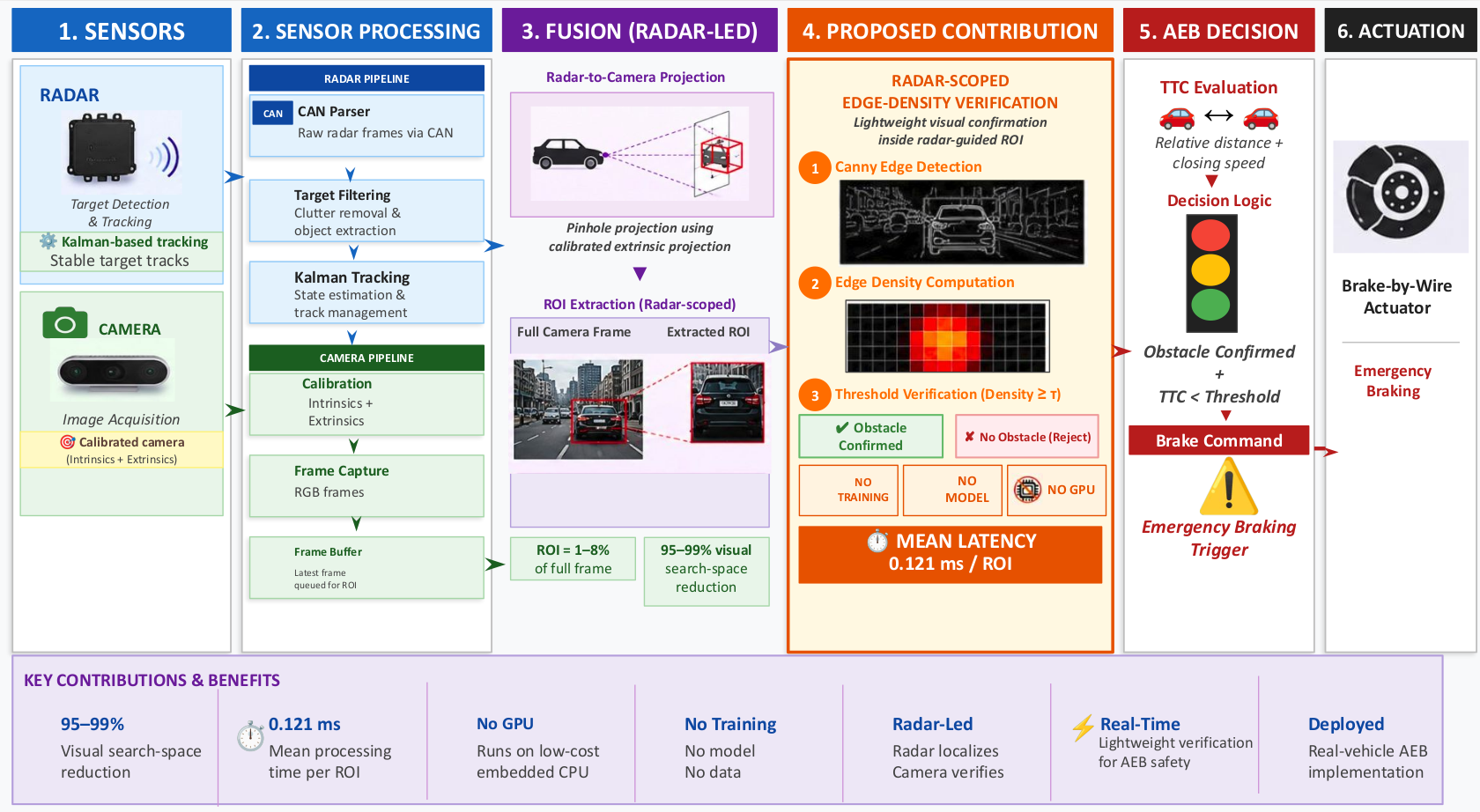}
\caption{Proposed radar--camera fusion AEB architecture. Radar detects and tracks obstacles, while the camera verifies obstacle presence within a radar-guided region of interest (ROI). Verified targets are combined with TTC estimates to trigger brake-by-wire actuation.}
\label{fig:architecture}
\end{figure*}

This work follows a different philosophy. Rather than improving object detection, once radar has already localized the target we examine whether object detection is required or not. In a radar-led AEB pipeline, localization is provided by the radar subsystem, allowing the system to process and operate only within a small radar-guided region of interest which is projected according to the relative distance. The camera therefore no longer needs to search the entire image and may not require full object recognition to support a braking decision. Instead, it only needs to determine whether an obstacle is present at the location identified by the radar. This changes the role of the camera from detecting things to verifying whether there is an obstacle or not and motivates the use of lightweight classical features that would be impractical in a full-frame perception pipeline. Table~\ref{tab:comparison} summarizes representative prior work and highlights the distinctions between existing fusion systems and the proposed architecture.

\begin{table*}[t]
\caption{Qualitative Comparison of Radar--Camera Fusion Systems Across AEB-Relevant Properties}
\label{tab:comparison}
\centering
\begin{tabular}{lccccc}
\hline
\textbf{System / Method} & \textbf{Learning} & \textbf{Real Vehicle} & \textbf{End-to-End AEB} & \textbf{Sub-ms Camera Stage} & \textbf{GPU-Free Deploy} \\
\hline
CenterFusion \cite{nabati2021centerfusion} & Yes & Not reported & No & No & Not reported \\
Zhou et al. \cite{zhou2023bridging} & Yes & Not reported & No & No & Not reported \\
CR3DT \cite{baumann2024cr3dt} & Yes & Not reported & No & No & Not reported \\
Lin et al. \cite{lin2023embeddedfusion} & Yes & Yes & No & No & No \\
Jahromi et al. \cite{jahromi2019hybridfusion} & Yes & Yes & No & No & No \\
\textbf{This work} & \textbf{No} & \textbf{Yes} & \textbf{Yes} & \textbf{Yes} & \textbf{Yes} \\
\hline
\end{tabular}
\end{table*}

\section{System Architecture}

\subsection{System Overview}

Our system was deployed on an instrumented Golf cart vehicle equipped with a Continental ARS408-21 radar \cite{continental2021ars408}, Intel RealSense D435 camera \cite{intel2020realsense}, and a custom brake-by-wire actuator with a chain sprocket design. Camera intrinsics and radar-to-camera extrinsics were calibrated to enable the projection of radar tracks into the image plane for ROI extraction and obstacle verification. Fig.\ref{test_vehicle}. show the sensor placement and test platform used throughout the experiments. All the staged tests were conducted in controlled environments with a safety driver present and manual override available at all times.

The overall architecture of the proposed radar-led AEB system is illustrated in Fig.\ref{fig:architecture}. The radar first detects and tracks potential obstacles and estimates their relative distance and velocity. These radar tracks are then projected onto the camera image to extract a small radar-guided region of interest, where the proposed edge-density gate performs obstacle verification. The verified obstacle and the radar-derived time-to-collision estimate are then used to trigger the brake-by-wire actuator.

\subsection{Radar Processing and Kalman Tracker}

ARS408 radar detections are sent over CAN and are parsed at the frame level \cite{continental2021ars408}. A pre-filter passes only forward-facing targets within 0--25~m with RCS above $-20$~dBsm. Nearby detections sharing similar velocities are centroid-merged to handle multiple returns from the same physical object. Merged detections feed a constant-velocity Kalman tracker \cite{kalman1960filter}:

\begin{equation}
x_k =
\begin{bmatrix}
x & y & v_x & v_y
\end{bmatrix}^{T}
\end{equation}

A track is promoted to confirmed status only after appearing in three consecutive radar updates, eliminating the vast majority of ghost returns. An optional velocity gate filters tracks moving away from the ego vehicle.

\subsection{Camera--Radar Projection and ROI Extraction}

Each confirmed radar track gets projected into the image via the standard pinhole model:

\begin{equation}
s
\begin{bmatrix}
u\\
v\\
1
\end{bmatrix}
=
K
\left[
R \mid t
\right]
\begin{bmatrix}
x\\
y\\
z\\
1
\end{bmatrix}
\end{equation}

where $z$ runs from 0 to 2.2~m. The projected pixels get enclosed in an axis-aligned bounding box with 22~px padding---at 8~m that's 4.3\% of the frame, at 25~m it shrinks to 1.3\%. Small windows.

\subsection{Edge-Density Gate}

Given ROI patch $P$ of dimensions $h \times w$ pixels, the system converts to grayscale and runs Canny with fixed thresholds---low = 40, high = 120 \cite{canny1986edge}. Edge pixels are then counted and normalised:

\begin{equation}
d(P)=\frac{1}{h \cdot w}\sum_{u,v}\mathbf{1}\!\left[E(P)(u,v)>0\right]
\end{equation}

where $E(P)$ is the binary Canny edge map. If $d(P)\geq\tau$, the radar track is confirmed. That is the entire camera stage.

The deployed threshold is $\tau=0.025$---at least 2.5\% of ROI pixels must be edges for the gate to pass. In practice, real obstacles at close range produce edge densities well above this; empty road and sky rarely do. One scalar threshold. No model weights, no training pipeline, no GPU at deploy time.

\subsection{AEB Decision Logic}

Time-to-collision and the brake condition:

\begin{equation}
TTC=\frac{r}{|v_r|}, \qquad
TTC_{\mathrm{eff}}=TTC-1.0~\mathrm{s}
\end{equation}

\begin{equation}
\mathrm{brake}=1
\quad\text{iff}\quad
d(P)\geq\tau
\;\land\;
r\in[3,25]~\mathrm{m}
\;\land\;
TTC_{\mathrm{eff}}<2.5~\mathrm{s}
\end{equation}

When the conditions in (4) are satisfied, a brake command will be sent to the BBW actuator over USB serial.

\begin{figure}[t]
\centering
\includegraphics[width=\columnwidth]{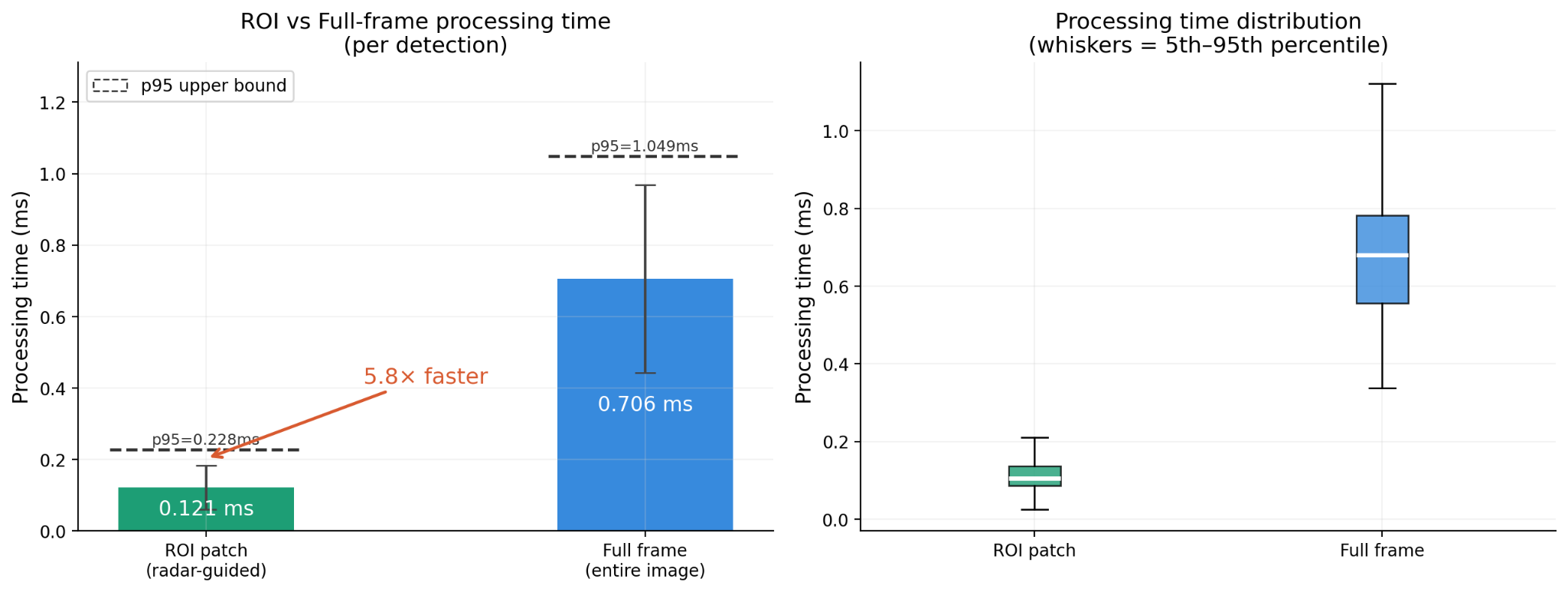}
\caption{ROI vs. full-frame processing time. Bars show mean; box plot shows 5th--95th percentile. ROI is 5.8$\times$ faster with p95 = 0.228~ms.}
\label{fig:latency}
\end{figure}

\section{Latency Evaluation}

\subsection{Measurement Setup}

Time was measured during execution of the Canny edge detector \cite{canny1986edge}. Two cases were evaluated mainly: processing only the radar-guided ROI and processing the entire $848 \times 480$ image. All tests were performed on the same laptop without GPU acceleration. The evaluation used data from 72 driving sessions and 49,988 frames, resulting in 41,452 individual ROI timing samples collected across daytime, evening, highway, and controlled driving scenarios.

\subsection{Results}

Fig.\ref{fig:latency} and Table~\ref{tab:latency} summarize the latency results. The mean processing time was 0.121 ms for ROI-based operation and 0.706 ms for full-frame processing, corresponding to a 5.8$\times$ reduction in latency. The 95th percentile latency remained below 0.228 ms for ROI processing, indicating stable performance across all recorded sessions.

\begin{table}[t]
\caption{Processing Time Comparison---ROI Patch vs. Full-Frame Canny}
\label{tab:latency}
\centering
\begin{tabular}{lccc}
\hline
\textbf{Metric} & \textbf{ROI Patch} & \textbf{Full Frame} & \textbf{Improvement} \\
\hline
Mean latency (ms) & 0.121 & 0.706 & 5.8$\times$ faster \\
Std. deviation (ms) & 0.061 & 0.263 & --- \\
p95 latency (ms) & 0.228 & 1.049 & 4.6$\times$ lower \\
Frames analysed & 49,988 & 49,287 & --- \\
Pixels at 8 m (\% of frame) & 4.3\% & 100\% & 95.7\% reduction \\
Pixels at 25 m (\% of frame) & 1.3\% & 100\% & 98.7\% reduction \\
Recording sessions & 72 & --- & --- \\
\hline
\end{tabular}
\end{table}

The reduction in processing time is mostly because of the smaller number of pixels contained within the radar-guided ROI. Across the evaluated operating range, the ROI occupied approximately 1--17\% of the full image, with smaller regions observed at longer target distances. As a result, the computational advantage increased as the target moved farther from the vehicle. Consistent latency behaviour was observed across all 72 driving sessions, including evening and dense traffic conditions.

\begin{figure*}[t]
\centering
\includegraphics[width=\textwidth]{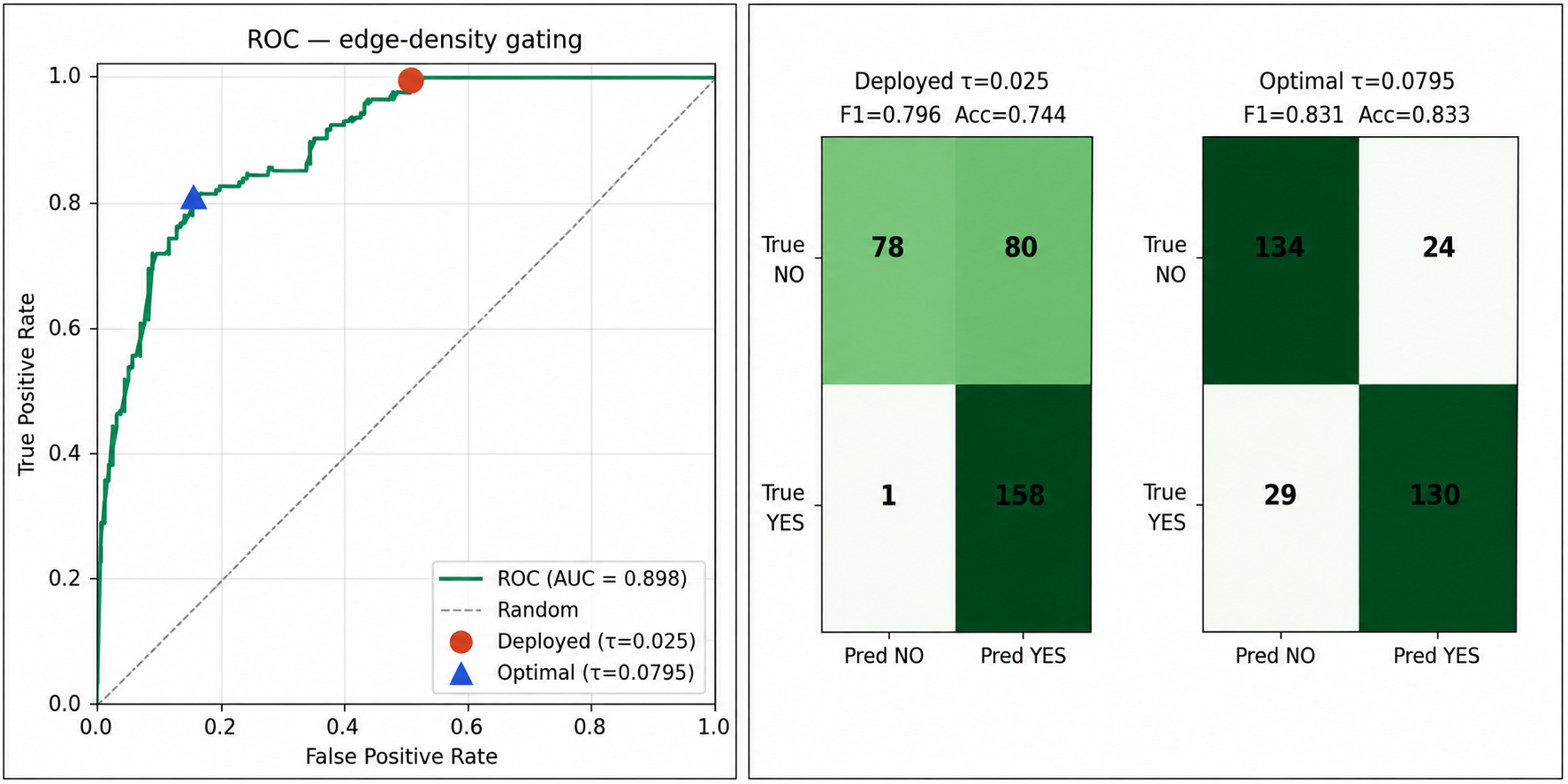}
\caption{Left: ROC curve---AUC = 0.898. Red marker: deployed threshold ($\tau=0.025$, recall = 0.994). Blue marker: Youden-optimal threshold ($\tau=0.080$, F1 = 0.831). Right: Confusion matrices at deployed threshold (left matrix) and Youden-optimal threshold (right matrix).}
\label{fig:roc_confusion}
\end{figure*}

\section{Gating Quality Evaluation}

\subsection{Labelling Protocol}

\begin{figure*}[t]
\centering
\includegraphics[width=\textwidth]{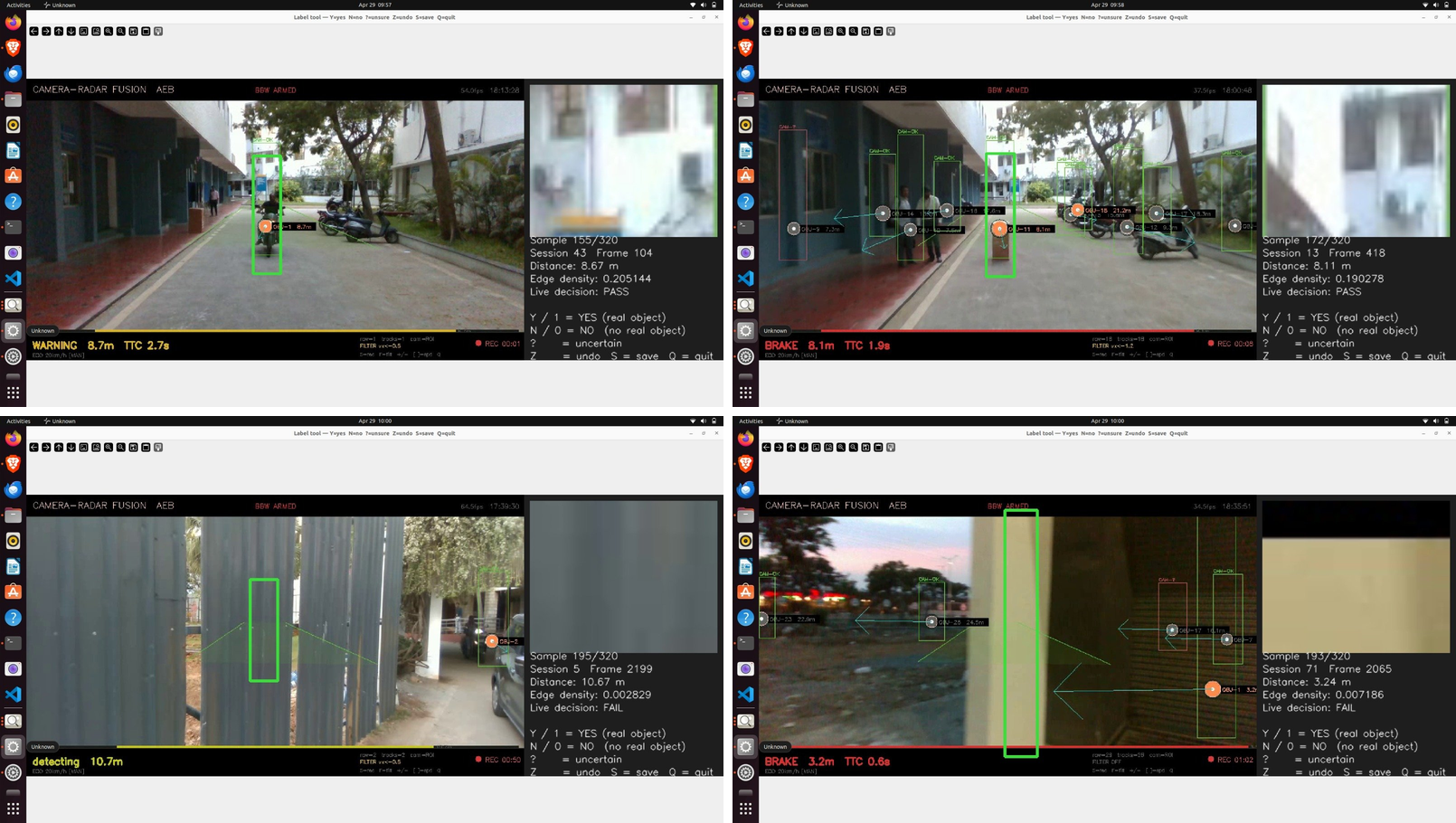}
\caption{Sample frames from the label tool. Top: motorcyclist at 8.7 m, $d=0.205$, PASS; pedestrian at 8.1 m, $d=0.190$, PASS. Bottom: corrugated metal sheet at 10.7 m, $d=0.003$, FAIL; building facade in near-darkness at 3.2 m, $d=0.007$, FAIL. The green box marks the active ROI; the right panel shows the ROI crop.}
\label{fig:sample_frames}
\end{figure*}

Edge density was computed offline for 60,492 radar-projected ROIs across all 72 sessions. From this pool, 320 frames were selected using stratified sampling across four density bands with at least 50\% from controlled scenario sessions. Each frame was labelled YES or NO by the first author. Three ambiguous frames were excluded, leaving 317 labels---159 YES, 158 NO. To reduce selection bias, frames were stratified across edge-density ranges and operating conditions before annotation. Representative examples of the labelled ROI patches used during annotation are shown in Fig.\ref{fig:sample_frames}. The figure includes both obstacle and non-obstacle cases together with their corresponding edge-density values.

\subsection{ROC Analysis and Threshold Selection}

AUC = 0.898 (Fig.~\ref{fig:roc_confusion}). The deployed threshold is $\tau = 0.025$. One object was missed out of 159---recall = 0.994. Increasing $\tau$ to 0.080 improves precision to 0.844 and F1 score to 0.831, but recall drops by 18 percentage points. Table~\ref{tab:gating_performance} summarizes the performance at both thresholds.

\begin{table}[t]
\caption{Gating Performance at Deployed and Youden-Optimal Thresholds ($n=317$)}
\label{tab:gating_performance}
\centering
\begin{tabular}{lcc}
\hline
\textbf{Metric} & \textbf{Deployed} ($\tau=0.025$) & \textbf{Youden-Optimal} ($\tau=0.080$)\\
\hline
True Positives & 158 & 130 \\
False Positives & 80 & 24 \\
True Negatives & 78 & 134 \\
False Negatives & 1 & 29 \\
Precision & 0.664 & 0.844 \\
Recall & 0.994 & 0.818 \\
F1 Score & 0.796 & 0.831 \\
Accuracy & 0.744 & 0.833 \\
AUC & 0.898 & 0.898 \\
\hline
\end{tabular}
\end{table}

\subsection{Threshold Sensitivity}

Table~\ref{tab:threshold_sensitivity} sweeps six thresholds. The F1 score peaks somewhere between $\tau=0.050$ and $\tau=0.080$. Below $\tau=0.025$ the gate lets too much through. Above $\tau=0.100$ recall collapses faster than precision gains anything useful---not acceptable for AEB.

\begin{table}[t]
\caption{Threshold Sensitivity---Precision, Recall, F1 Across Candidate Values}
\label{tab:threshold_sensitivity}
\centering
\begin{tabular}{lcccc}
\hline
\textbf{Threshold} & \textbf{Precision} & \textbf{Recall} & \textbf{F1} & \textbf{Notes}\\
\hline
0.010 & 0.526 & 1.000 & 0.690 & Too permissive \\
0.025 & 0.664 & 0.994 & 0.796 & Deployed---AEB-safe \\
0.050 & 0.829 & 0.824 & 0.826 & Balanced operation \\
0.080 & 0.844 & 0.818 & 0.831 & Youden-optimal \\
0.100 & 0.839 & 0.786 & 0.812 & Conservative \\
0.150 & 0.937 & 0.465 & 0.622 & Unsafe---too many misses \\
\hline
\end{tabular}
\end{table}

\subsection{Illumination-Stratified Breakdown}

Table~\ref{tab:illumination} shows the results across daytime controlled, daytime naturalistic, and evening conditions. Recall remained 1.000 in all three groups. Precision decreased during evening operation due to additional edge responses from street lighting and reflective surfaces. While false confirmations increased, obstacle confirmation performance remained consistent across all evaluated conditions.

\begin{table}[t]
\caption{Performance Across Different Lighting Conditions at the Deployed Threshold ($\tau=0.025$)}
\label{tab:illumination}
\centering
\begin{tabular}{lcccc}
\hline
\textbf{Condition} & \textbf{n} & \textbf{Precision} & \textbf{Recall} & \textbf{F1}\\
\hline
Controlled daytime (sessions 27--63) & 99 & 0.768 & 1.000 & 0.869 \\
Daytime naturalistic (sessions 0--26) & 126 & 0.779 & 0.988 & 0.871 \\
Evening / low-light (sessions 64--71) & 92 & 0.269 & 1.000 & 0.424 \\
All sessions combined & 317 & 0.664 & 0.994 & 0.796 \\
\hline
\end{tabular}
\end{table}

\section{AEB Behavioural Evaluation}
\subsection {Scenario Design}
\begin{figure*}[t]
\centering
\includegraphics[width=\textwidth]{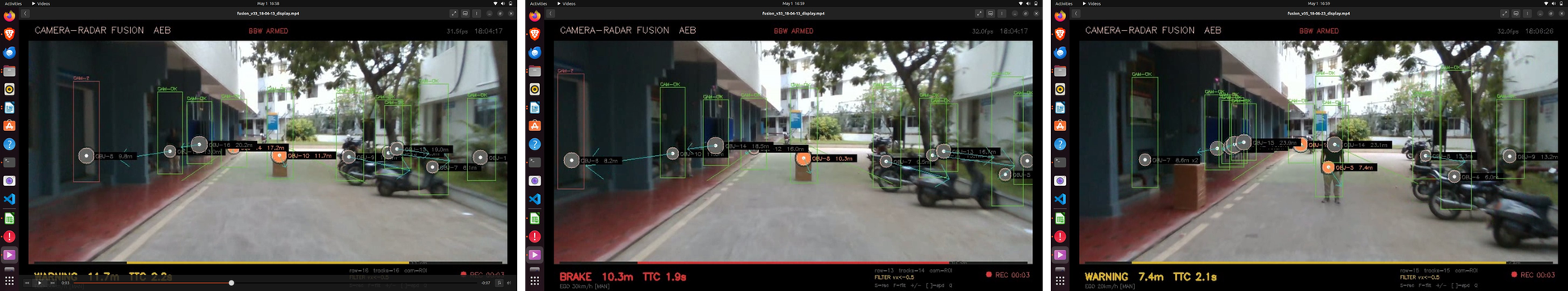}
\caption{Staged scenario camera feed. Left: cardboard box at approximately 11 m in the WARNING state (18:04 hrs). Centre: cardboard box at 10.3 m with BRAKE triggered during a 30 km/h approach. Right: static pedestrian at 7.4 m in the WARNING state during a 20 km/h approach. The images illustrate the radar-guided ROI verification and corresponding AEB decision states under representative test scenarios.}
\label{fig:staged_scenarios}
\end{figure*}

The evaluation included 33 threat scenarios and 3 no-threat scenarios which we carefully created with our team members . The threat scenarios covered stationary obstacles, pedestrians, bicycles, crossing pedestrians, and closing-velocity situations with different properties. Vehicle speeds ranged from 20 km/h to 30 km/h depending on the scenario.
We also included challenging situations , such as pedestrians emerging from behind parked vehicles and multi-object scenarios involving a pedestrian, bicycle, and cardboard box. Multiple runs were conducted for each scenario to verify consistent system behaviour. Representative examples are shown in Fig. \ref{fig:staged_scenarios}.
The no-threat scenarios were mostly objects located near the vehicle but outside the collision path. These tests were used to observe the behaviour of the system when braking was not required. The staged scenarios were designed following the Euro NCAP AEB Car-to-Car Test Protocol \cite{euroncap2022aeb}. The staged scenario camera feed is shown in Fig. \ref{fig:staged_scenarios}.

\subsection {Results}
\begin{figure*}[t]
\centering
\includegraphics[width=\textwidth]{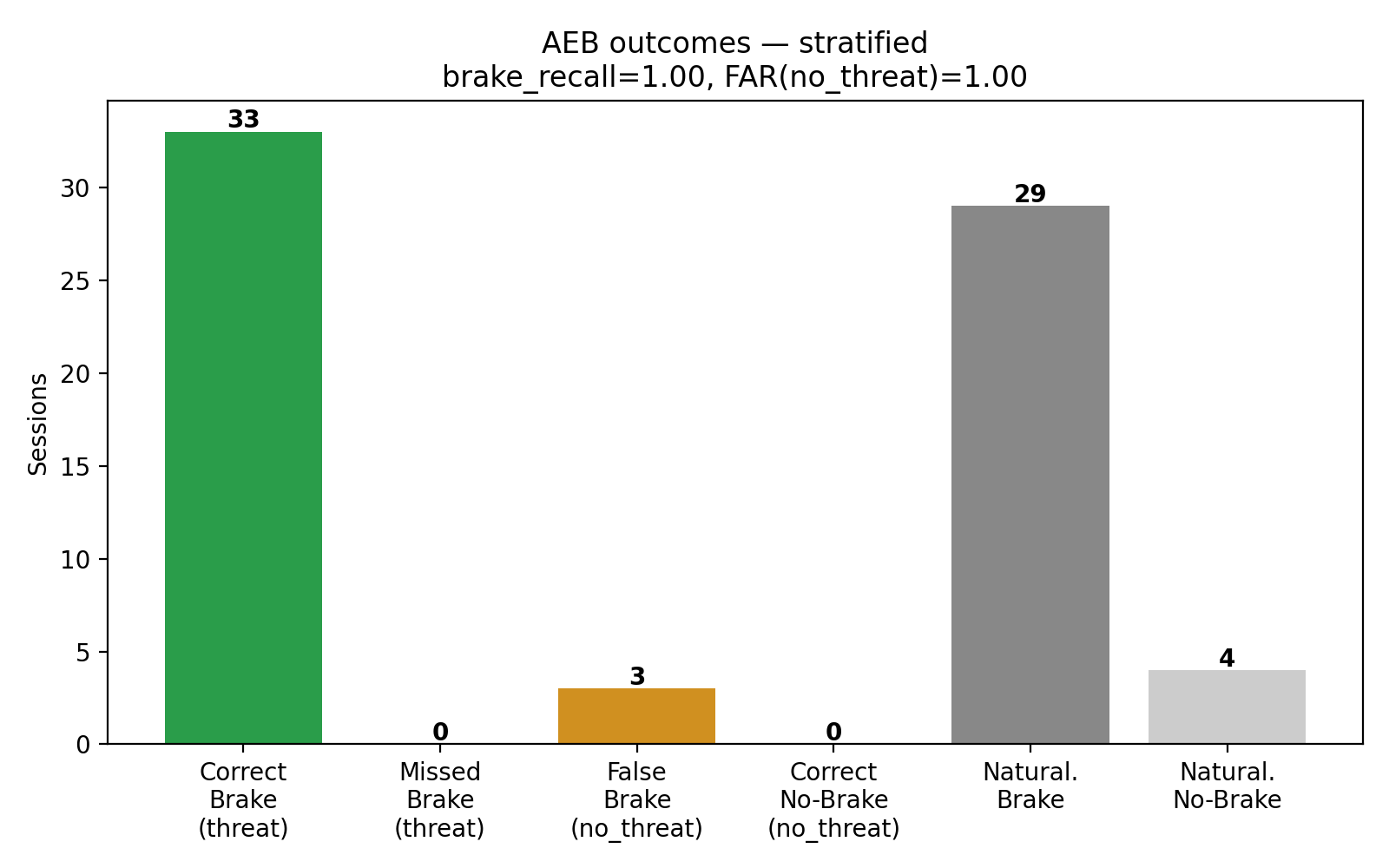}
\caption{AEB outcomes stratified by scenario type. Brake recall = 1.000 across all staged threat scenarios. False alarm rate = 1.000 on three deliberate no-threat sessions, consistent with the deployed threshold ($\tau = 0.025$). Naturalistic brake rate = 0.879 (29/33 sessions), with all brake events corresponding to genuine in-corridor obstacles.}
\label{fig:aeb_outcomes}
\end{figure*}

\begin{figure*}[t]
\centering
\includegraphics[width=\textwidth]{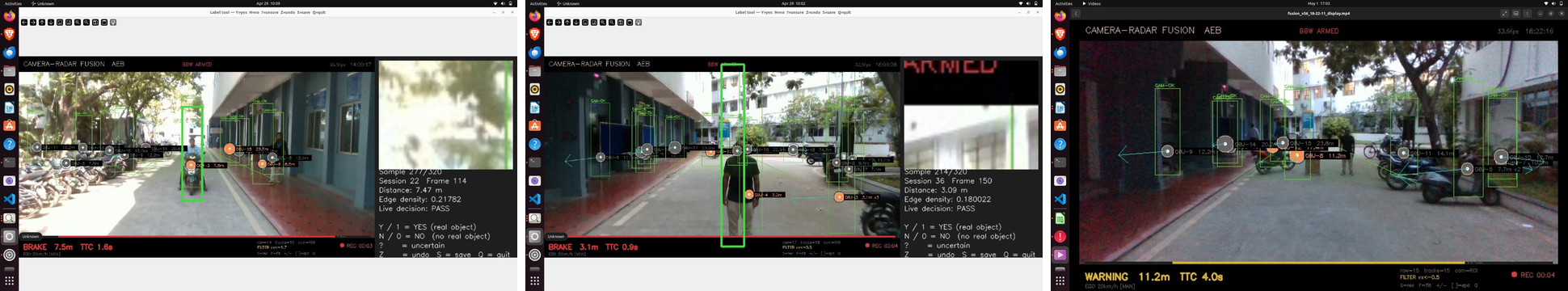}
\caption{System camera feeds across representative operating conditions. Left: daytime dense naturalistic traffic with multiple confirmed radar-guided tracks. Centre: pedestrian at 3.09~m in the BRAKE state with TTC = 0.9~s and edge density = 0.180. Right: evening low-light operation with a target at 11.2~m in the WARNING state (TTC = 4.0~s), demonstrating radar-guided obstacle verification under reduced illumination.}
\label{fig:operating_conditions}
\end{figure*}

The overall behavioural evaluation results are summarized in \ref{fig:aeb_outcomes}. Brake recall = 1.000 across all 33 staged threat sessions --- zero missed brakes. In all 3 no-threat sessions a brake event occurred --- strict FAR = 1.000 on this category, consistent with the high-recall operating point. Raising $\tau$ to 0.080 would be expected to substantially reduce false triggers at a recall cost the operator must decide is acceptable. In 33 naturalistic sessions, brakes triggered in 29; manual review confirmed a genuine in-corridor obstacle in every triggered event. The 4 sessions with no brake had no qualifying close-range obstacles during the recording window as illustrated in Fig. \ref{fig:aeb_outcomes}. The algorithm is also tested in different lighting conditions as shown in Fig. \ref{fig:operating_conditions}.

\section{Discussion}

\subsection{Verification Rather Than Detection}

Our proposed gate achieved a recall of 0.994 and an AUC of 0.898 while operating on small radar-guided regions of interest. Across the 33 staged threat scenarios, the system recorded no missed brake events. These results suggest that, for radar-led AEB, obstacle verification may be sufficient once radar has already localized the target. Rather than searching the entire scene and identifying objects, the camera only needs to confirm whether an obstacle is present at the radar-indicated location.

\subsection{Deployment Implications}

The camera verification stage achieved a mean latency of 0.121 ms per ROI while reducing the visual search space by approximately 95--99\%. The system operates without training data, model weights, or dedicated AI hardware. These results indicate that lightweight camera verification can provide a practical alternative to detector-based confirmation in resource-constrained AEB systems.

\section{Limitations and Future Work}
Our current system has some limitations. The largest limitation is radar--camera calibration accuracy. Small projection errors become more noticeable at shorter distances and can affect ROI placement. The dataset was annotated by a single reviewer. A larger multi-annotator study with more test data would help us to  measure annotation consistency. The experiments were conducted using a single vehicle platform where we have configured the sensors. So additional testing across different vehicles, environments, and lighting conditions is still required for better understanding .In the results we can see that reduction in  precision during evening operation. This suggests that a fixed edge-density threshold may not be optimal under all lighting conditions.
Future work will focus on improving calibration accuracy and increasing robustness to changing illumination. The current implementation was developed in Python using OpenCV, and additional performance gains may be possible through a more optimized implementation. Monochrome cameras are another area of interest because they can provide stronger edge contrast while maintaining the lightweight verification approach used in this work.

\section{Conclusion}

Our work was motivated by two simple questions. If radar can detect  and localize a target, does the camera still need to process the entire image? And if the objective is emergency braking, does the camera need to identify the obstacle before making a decision  for  braking ?
Based on the results that we achieved, both questions are worth reconsidering. Once the target location is available from radar, camera processing can be restricted to a small region of interest rather than the full image. Likewise, for AEB, obstacle verification may be sufficient in many situations without requiring full object recognition.
The observation from this work is not that perception systems should become simpler in every application. Rather, it is that the requirements of the task should drive the complexity of the solution. For AEB, the priority is making a timely and reliable braking decision. The results presented here suggest that this direction is worth exploring further, particularly for resource-constrained systems where computational efficiency is as important as perception performance.

\bibliographystyle{IEEEtran}
\bibliography{references}

\vspace{1em}

\begin{IEEEbiography}
[{\includegraphics[width=1.1in,height=2.0in,clip,keepaspectratio]{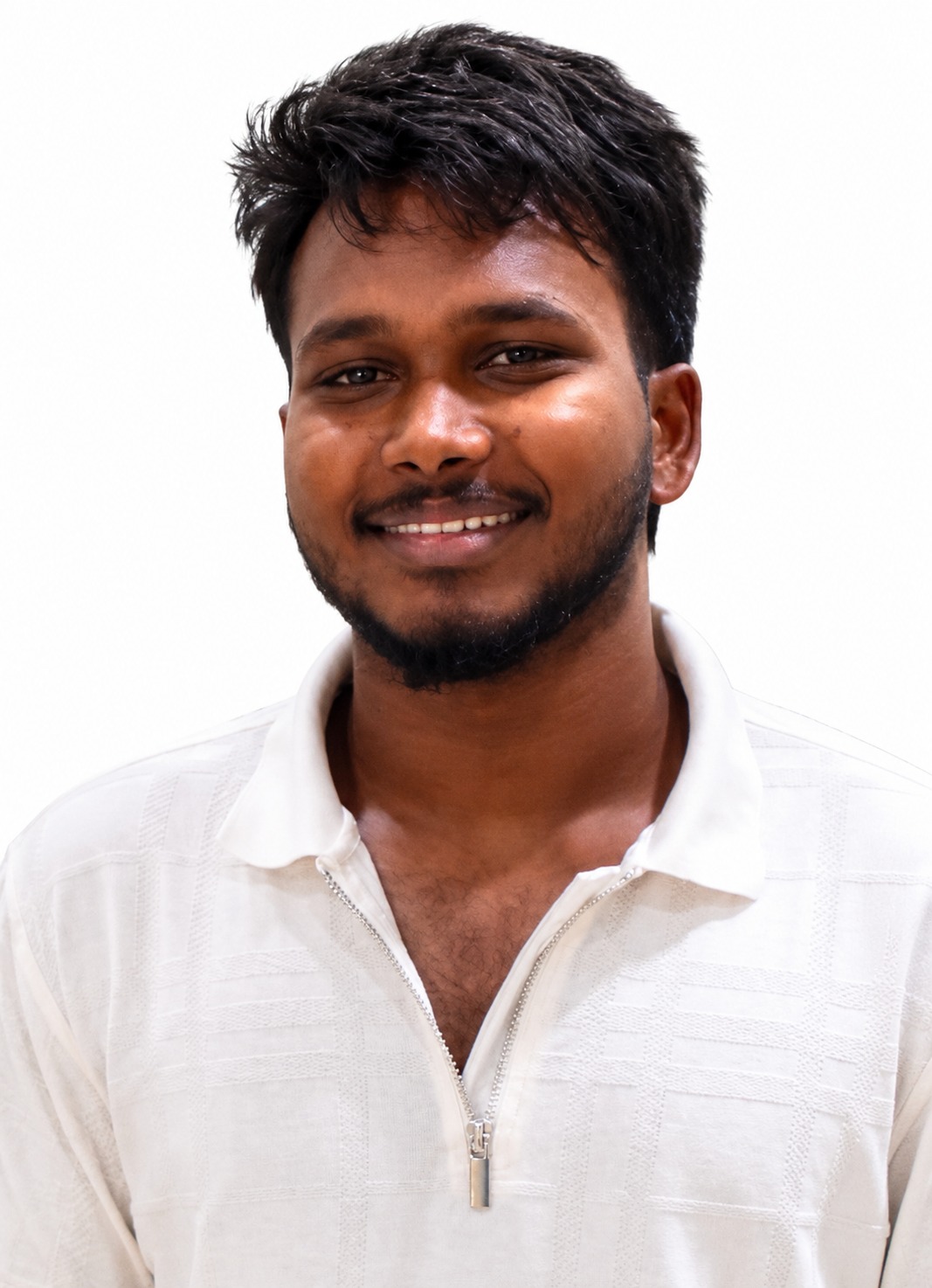}}]{RAM CHARAN AKULA}
is currently pursuing the B.Tech. degree in Mechatronics Engineering (Robotics) at SRM Institute of Science and Technology, Chennai, India. He is an intern at Euler Motors, New Delhi, India. As a core member of Electruisers, he has worked on autonomous driving and sensor-fusion systems for student-built vehicles. His interests include autonomous vehicles, ADAS, radar--camera fusion, and embedded perception.
\end{IEEEbiography}
\vspace{1em}

\begin{IEEEbiography}[{\includegraphics[width=1.1in,height=2.0in,clip,keepaspectratio]{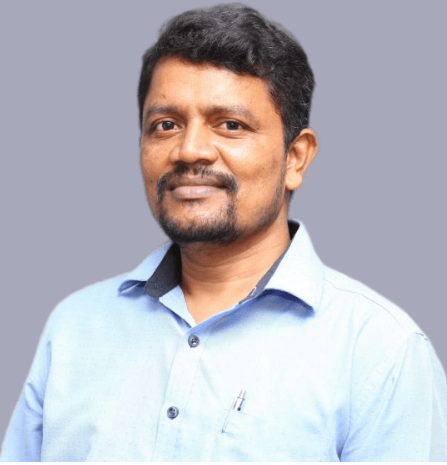}}]{Sivanathan Kandhasamy}
received the B.E. degree in Electronics and Communication Engineering, the M.E. degree in Mechatronics Engineering and the Ph.D. degree in robotics and autonomous systems. He is currently an Associate Professor with the Department of Mechatronics Engineering, SRM Institute of Science and Technology, Chennai, India, where he heads the Autonomous Systems Laboratory. His research interests include autonomous vehicles, intelligent transportation systems, robotics, artificial intelligence, perception, and sensor fusion. He has authored numerous journal articles, conference papers, patents, and has led several externally funded research and industrial consultancy projects in autonomous systems.
\end{IEEEbiography}
\vspace{1em}

\begin{IEEEbiography}
[{\includegraphics[width=1.1in,height=2.0in,clip,keepaspectratio]{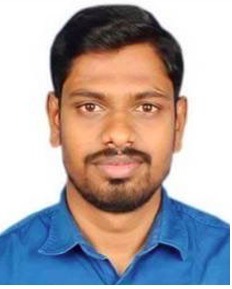}}]{Manikandan Ganesan}
received the B.E. degree in Electrical and Electronics Engineering and the M.E. degree in Mechatronics Engineering from Anna University, Chennai, India. He is currently a Project Associate and Research Scholar with the Center for Electric Mobility, SRM Institute of Science and Technology, Chennai, India. His research interests include autonomous vehicles, deep learning, motion planning, and mechatronic systems.
\end{IEEEbiography}

\end{document}